\documentclass[conference]{IEEEtran}
\IEEEoverridecommandlockouts

\usepackage{cite}
\usepackage{amsfonts}
\usepackage{amsmath}
\usepackage{subcaption} 
\usepackage{keyval}
\usepackage{algorithmic}
\usepackage{amssymb}
\usepackage{graphicx}
\usepackage{comment}
\usepackage{caption}
\usepackage{tcolorbox}
\usepackage{alltt}
\usepackage[hyphens]{url}
\usepackage{textcomp}
\usepackage{amsthm} 
\usepackage{listings}
\usepackage{tabularx}
\usepackage{booktabs}
\usepackage{array}
\usepackage{multirow}
\usepackage[hyphens]{url}
\usepackage{pifont}
\usepackage{hyperref}
\usepackage{pythonhighlight}
\usepackage{xcolor}

\definecolor{promptColor}{HTML}{1F77B4}   
\definecolor{guidelineColor}{HTML}{FF7F0E} 
\definecolor{textColor}{HTML}{2CA02C}      
\definecolor{codegreen}{rgb}{0,0.6,0}    
\definecolor{codeblue}{rgb}{0.2,0.2,0.8}  
\definecolor{codepurple}{rgb}{0.58,0,0.82} 
\definecolor{codegray}{rgb}{0.5,0.5,0.5}  
\definecolor{backcolour}{rgb}{0.95,0.95,0.92} 

\def\BibTeX{{\rm B\kern-.05em{\sc i\kern-.025em b}\kern-.08em
    T\kern-.1667em\lower.7ex\hbox{E}\kern-.125emX}}

\begin{document}


\title{Advanced Tool Learning and Selection
System (ATLASS): A Closed-Loop Framework Using LLM}

\author{
\IEEEauthorblockN{Mohd Ariful Haque\textsuperscript{1}, Justin Williams\textsuperscript{2}, Sunzida Siddique\textsuperscript{3}}\\
\vspace{-0.4cm}  
\IEEEauthorblockN{Md. Hujaifa Islam\textsuperscript{4}, Hasmot Ali\textsuperscript{5}, Kishor Datta Gupta\textsuperscript{6}, Roy George\textsuperscript{7}}
\IEEEauthorblockA{
Dept. of Cyber Physical Systems\textsuperscript{1,2,6,7}, Clark Atlanta University\textsuperscript{1,2,6,7}, USA\\
Dept. of Computer Science and Engineering\textsuperscript{3,4,5}, Daffodil International University, Bangladesh\textsuperscript{3,5}\\
Ahsanullah University of Science and Technology\textsuperscript{4}\\
\texttt{mohdariful.haque@students.cau.edu\textsuperscript{1}, sunzida15-9667@diu.edu.bd\textsuperscript{3}, hujaifaislam123@gmail.com\textsuperscript{4}}\\
\texttt{hasmot15-9632@diu.edu.bd\textsuperscript{5}, kgupta@cau.edu\textsuperscript{6}, rgeorge@cau.edu\textsuperscript{7}}
}
}

\maketitle

\begin{abstract}
The combination of LLM agents with external tools enables models to solve complex tasks beyond their knowledge base. Human-designed tools are inflexible and restricted to solutions within the scope of pre-existing tools created by experts. To address this problem, we propose ATLASS, an advanced tool learning and selection system designed as a closed-loop framework. It enables the LLM to solve problems by dynamically generating external tools on demand. In this framework, agents play a crucial role in orchestrating tool selection, execution, and refinement, ensuring adaptive problem-solving capabilities. The operation of ATLASS follows three phases: The first phase, Understanding Tool Requirements, involves the Agents determining whether tools are required and specifying their functionality; the second phase, Tool Retrieval/Generation, involves the Agents retrieving or generating tools based on their availability; and the third phase, Task Solving, involves combining all the component tools necessary to complete the initial task. The Tool Dataset stores the generated tools, ensuring reusability and minimizing inference cost. Current LLM-based tool generation systems have difficulty creating complex tools that need APIs or external packages. In ATLASS, we solve the problem by automatically setting up the environment, fetching relevant API documentation online, and using a Python interpreter to create a reliable, versatile tool that works in a wider range of situations. OpenAI GPT-4.0 is used as the LLM agent, and safety and ethical concerns are handled through human feedback before executing generated code. By addressing the limitations of predefined toolsets and enhancing adaptability, ATLASS serves as a real-world solution that empowers users with dynamically generated tools for complex problem-solving.

\end{abstract}


\section{Introduction}

Large Language Models (LLMs) have capabilities in a wide range of tasks, from natural language processing to content generation and problem solving \cite{wang-etal-2023-self-instruct, minaee2024largelanguagemodelssurvey}, including support for user input from all domains and multimodal ability \cite{zhang-etal-2024-mm}. These attributes enable user demands to be met in different applications, including chat systems, intelligent virtual agents, automated content generation, and code synthesis \cite{zhao2024surveylargelanguagemodels, brown2020languagemodelsfewshotlearners}. Although LLMs have achieved remarkable performance broadly, they still have inherent limitations, such as out-of-date information \cite{10.1145/3641289} and suffer from performance degradation \cite{ahmed2024studyingllmperformanceclosed}. Addressing these problems requires overcoming the limitations of pre-defined pipelines, which have restricted flexibility to calibrate incorrect actions. Additionally, it is challenging to adapt a general LLM-based agent to handle a wide range of specialized tasks. \cite{shi2024learningusetoolscooperative}. LLM agents are autonomous systems that combine reasoning, perception, and action to perform tasks. They bridge general-purpose LLMs and domain-specific needs by structuring complex problems into step-by-step reasoning. Researchers have introduced collaborative environments where multiple intelligent agent components, each with distinctive attributes and roles, work together to handle complex tasks more efficiently and effectively \cite{talebirad2023multiagentcollaborationharnessingpower}. 

LLM agents often struggle with complex tasks that require external knowledge, computations, or real-world interactions beyond their pretrained capabilities. Tool learning, the ability of LLM agents to use external resources, is used to help LLM agents with various auxiliary resources, like search engines \cite{ nakano2022webgptbrowserassistedquestionansweringhuman, qin-etal-2023-webcpm} or calculators \cite{gao2023palprogramaidedlanguagemodels, schick2023toolformerlanguagemodelsteach} which empower them as tool-user agents and improve their ability to tackle complex tasks. Tool-based agents generally work by breaking down a task and planning a sequence of tools to complete it step by step. For each step, the agent executes the tools by passing arguments and continuously incorporating useful intermediates into the next action prediction. However, this approach has difficulty in  adapting a single LLM-based agent to learn multiple specialized actions in solving a task. This limitation reflects a broader challenge:  smaller models struggle with complex tasks, while specialized agents better handle tool selection and execution. To mitigate this problem, ConAgents coordinates three specialized agents for tool selection, tool execution, and action calibration separately \cite{shi2024learningusetoolscooperative}. 

In this work, we present the ATLASS  framework to automate tool selection, generation, and execution. The framework uses the generated tools for a single inference and stores them in the tools database for further use. 
There are three key aspects to this contribution:
\begin{itemize}
    \item \textbf{Understanding Tool Requirements:} Analyzing user prompts, defining subtasks, defining the need for tools, and identifying the appropriate tools from the toolset.
    
    \item \textbf{Tool Retrieval or Generation:} Retrieving and registering tools from the toolset or generating tools and registering them to the agent.
    
    \item \textbf{Task Solving:} Solving user tasks by using tools.
\end{itemize}

This approach requires detailed task analysis, tool requirement understanding, and the integration of a comprehensive tool generation module. Simpler approaches, like the one used by Large Language Models as Tool Makers (LATM) \cite{cai2024largelanguagemodelstool}, result in basic, task-specific tools that overlook the reuse of similar task requirements. These approaches also do not create generalized tools that permit this reusability. ATLASS addresses this limitation by analyzing user queries, breaking down the tasks, understanding tool requirements, and identifying that a single tool can efficiently handle similar queries. This approach creates reusable tools that may be applied to future tasks to reduce redundancy.

\section{Literature Review}

The development of large language models (LLMs) has been driven by significant advancements in pre-training, fine-tuning, and evaluation techniques. Researchers have explored different methodologies to enhance LLM performance, focusing on dataset quality, tuning strategies, and evaluation metrics.

Zhao et al. \cite{zhao2023survey} review LLM advancements, covering pre-training, tuning, and evaluation. They discuss important datasets like CommonCrawl, C4, and Wikipedia, and tuning methods such as FLAN and RLHF. They show that instruction tuning improves LLaMA models, and Chinchilla scaling improves parameter-to-data ratios. Beyond scaling and tuning, prompt engineering is crucial for LLM performance. Marvin et al. \cite{marvin2023prompt} explore techniques like few-shot learning, chain-of-thought prompting, and automatic instruction generation, finding that automated prompts outperform human-designed ones in 19 of 24 NLP tasks. However, hallucination, scalability, and bias remain challenges.

Wang et al. \cite{wang-etal-2023-self-instruct} introduce the self-instruct framework using GPT-3. Human evaluations indicate that it outperforms publicly trained models and performs comparably to InstructGPT-001. Liu et al. \cite{liu2024tuning} explore fine-tuning techniques, which enhance LLM adoption by enabling domain-specific customization and improving accuracy for specialized tasks. Hu et al. \cite{hu2022lora} propose parameter-efficient methods such as LoRA, which reduce costs and democratize access to LLMs. Minaee et al. \cite{minaee2024largelanguagemodelssurvey} discuss LLM adaptability, highlighting its role in expanding real-world applications and improving versatility across industries.

Duetting et al. \cite{duetting2025multi} review combinatorial contract theory in multi-agent settings with multiple actions, introducing a constant-factor approximation algorithm for submodular rewards and an FPTAS for single-agent cases. In parallel, Guo et al. \cite{guo2024large} review LLM-based multi-agent systems, focusing on agent-environment interaction, profiling, communication, and capability development. Their study covers applications in software development, embodied AI, and simulations, while also addressing challenges such as hallucination and scalability.

Xu et al. \cite{xu2025largereasoningmodelssurvey} review LE-MCTS, which employs process-level engineering to enhance reasoning paths, resulting in a 3–4\% improvement in math problem-solving accuracy. The AvaTaR framework \cite{wu2024avataroptimizingllmagents} focuses on optimizing LLM agents for tool use in complex tasks. The study reports a 15.6\% improvement in Hit@1 on the STARK benchmark and a 53\% exact match on HotpotQA.

A collaborative framework for tool learning is presented by Shi et al. \cite{shi2024learning}, where they utilize three specialized agents: grounding, execution, and review, communicating through automatic and adaptive protocols. Their Specialized Action Distillation (SPAN) technique adapts task-solving strategies from models like GPT-4 to open-source models, achieving a 14\% improvement in success rates on ToolBench and RestBench datasets. Similarly, the AutoAgents framework \cite{chen2023autoagents} dynamically creates task-specific agents using self-refinement and collaborative refinement. It improves task execution by incorporating observers to monitor the process. When tested on open-ended question answering and writing, it outperforms GPT-4.

Cai et al. \cite{cai2024largelanguagemodelstool} introduce the LATM framework for tool creation using language models, where training, validation, and test sets are generated based on user prompts. First, the framework generates a validation set and then uses the prompt to create a Python function to solve the task and finally validates the function before applying it to the test set. However, their method does not incorporate APIs for external information retrieval, limiting real-time adaptability. Our work addresses this gap by integrating SerpAPI \cite{serpapiSerpApiGoogle} for web-based search and retrieval, enhancing tool performance through dynamic, context-aware information extraction.

Google's MASS framework \cite{zhou} optimizes agent prompts, communication structures, and system instructions, streamlining multi-agent interactions. Meanwhile, Themis \cite{hu} fine-tunes LLMs for nuanced, context-aware evaluations, closely aligning with human judgments while revealing challenges in knowledge distillation. Additionally, Agentic reasoning \cite{wu} integrates external tools for tasks like web search and knowledge graph construction, excelling in scientific reasoning. These advancements enhance scalability, efficiency, and expertise in AI-driven systems.

Several researchers \cite{ni2025toolfactory} have worked on automated tool generation for scientific APIs, but real-world adoption is still hindered by inconsistent documentation, authentication, and access control. They propose that a combination of APILlama and a validation pipeline could improve the accessibility of AI agent development in research fields. Zhang et al. \cite{zhang2022automatic} introduce Auto-CoT, an approach for automating Chain-of-Thought (CoT) prompting in LLMs. Auto-CoT uses a clustering-based method to find diverse and representative examples, which reduces the amount of work that needs to be done by hand while maintaining strong reasoning abilities. Their approach matches or surpasses manually designed CoT prompting in math, commonsense, and symbolic reasoning tasks. However, LLM unreliability, potential error propagation, and high computational overhead present limitations to this method.

In this work, we've presented ATLASS, a framework tackling the limitations LLM Agents face when creating tools for complex tasks. ATLASS analyzes tasks, pulls relevant tools from existing sets, and builds new ones when needed. By adding real-time web search capabilities, our system adapts to changing contexts. The system's knack for breaking down problems and picking or making the right tools pushes autonomous agents forward while working well across different domains.

\section{ATLASS}
\begin{figure*}[!ht]
    \centering
    \includegraphics[width=\textwidth]{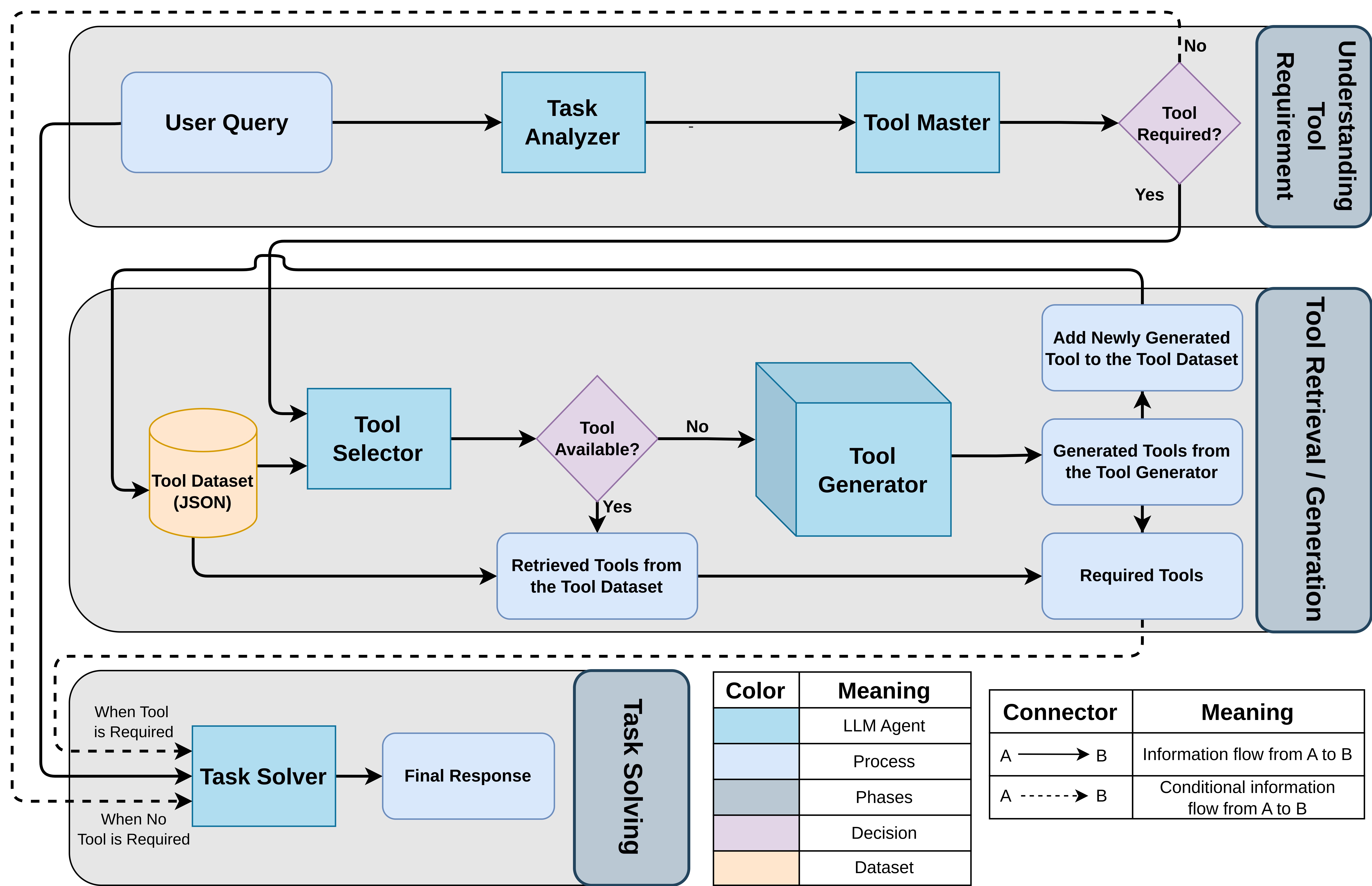}
 \caption{Overview of ATLASS workflow, with Tool Requirements Analysis, Tool Retrieval/Generation, and Task Execution.}
    \label{fig:fullpipeline}
\end{figure*}

ATLASS has a closed-loop architecture with multiple LLM agents, a multi-agent LLM, and a tool database. It can analyze user queries, break them down into subtasks, understand tool requirements, find existing tools and/or generate required tools, and solve user tasks using the generated or retrieved tools. Three key processes divide the ATLASS framework, each focusing on a distinct aspect of the overall workflow. Figure~\ref{fig:fullpipeline} shows the proposed ATLASS framework. 

\subsection{Tool Requirement Analysis}
This stage of the framework processes the user's initial query and determines whether an external tool is necessary to complete the task. It includes agents that rely only on the LLM's internal knowledge base to generate an appropriate response, without utilizing any external tools.

\begin{tcolorbox}[colback=gray!10, colframe=white, title=Example]
\small
\textbf{User Query: }\textit{Retrieve a list of the top 100 scientific books and organize them in ascending order.}
\end{tcolorbox}

\subsubsection{Task Analyzer}
\begin{figure}[!ht]
    \centering
    \includegraphics[width=0.5\textwidth]{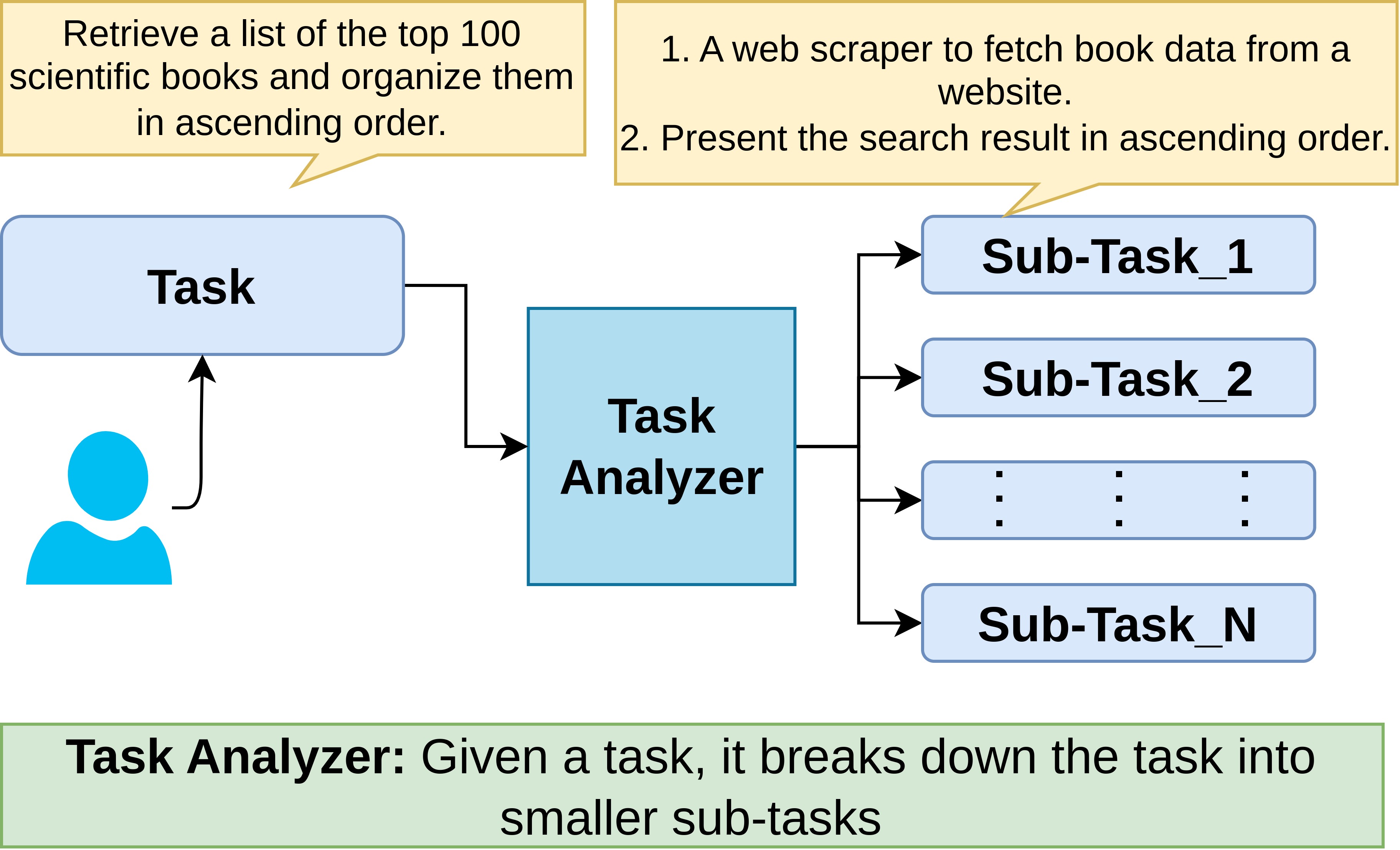}
    \caption{Task Analyzer Pipeline}
    \label{fig:taskanalyzer}
\end{figure}

The task analyzer takes the user's task and breaks it into multiple subtasks. This strategy enables the later agents to solve the problem step-by-step. The sub-tasks also help the tool master realize that an external tool may be required to do some of these sub-tasks. Figure~\ref{fig:taskanalyzer} demonstrates how Task Analyzer works. 

\begin{tcolorbox}[colback=gray!10, colframe=white, title=Example]
\small
\textbf{Task Analyzer:} \textit{\\1. A web scraper to fetch book data from a website.\\
2. Present the search result in ascending order.}
\end{tcolorbox}

\subsubsection{Tool Master}
\begin{figure}[!ht]
    \centering
    \includegraphics[width=0.5\textwidth]{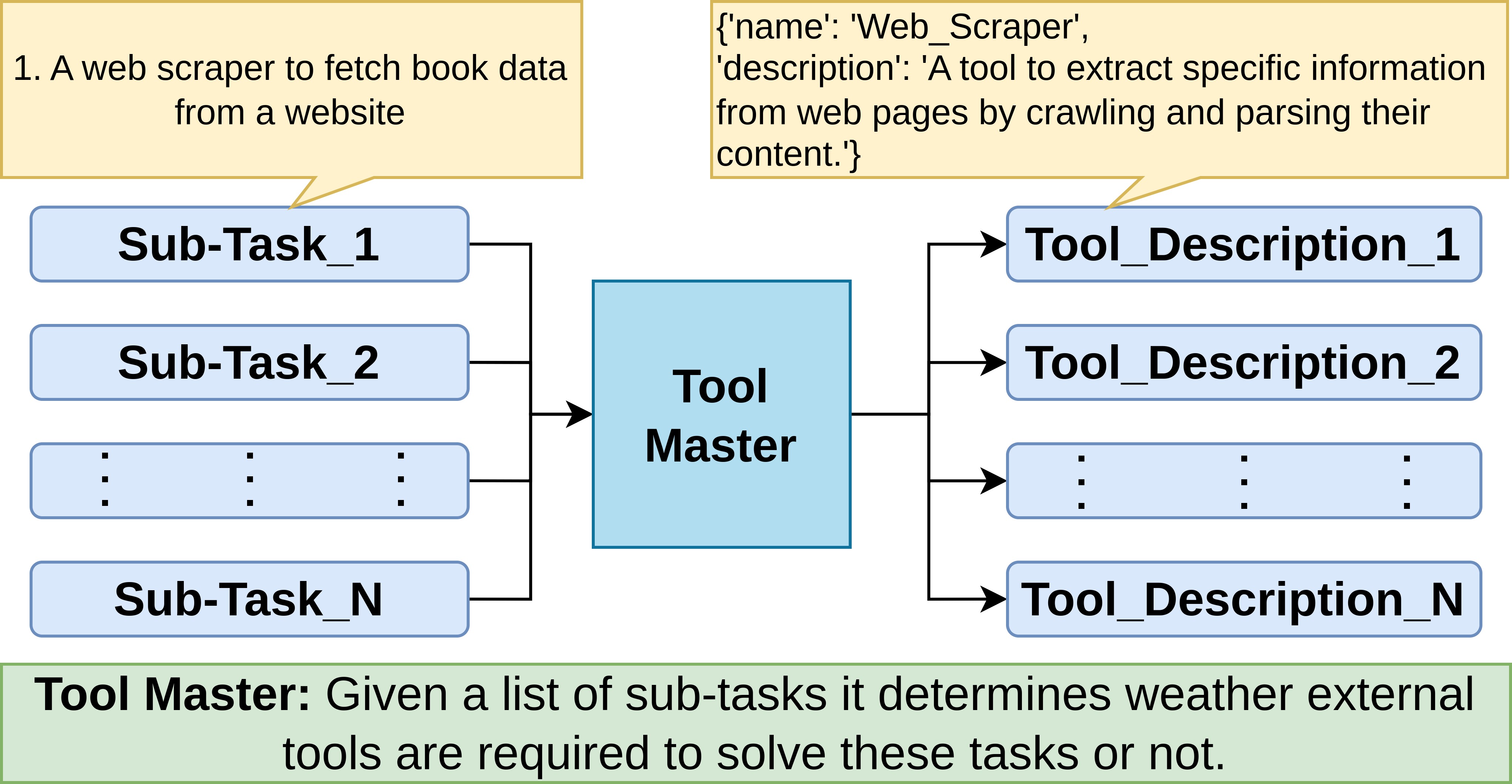}
    \caption{Tool Master Pipeline}
    \label{fig:toolmaster}
\end{figure}

An agent takes the breakdown of the sub-tasks from the Task Analyzer and determines whether external tools are required to solve the task or not. If external tools are required, this agent provides the 'name' and 'description' of all the required tools in a JSON format. However, if no tool is required, this agent simply responds that it doesn't require any tool. Figure \ref{fig:toolmaster} shows the pipeline for the Tool Master agent.

If the tool master determines that an external tool is needed, then we need to retrieve or generate it using the second stage of the pipeline. Otherwise, the user query is directly sent to the final stage of the framework, where the agent solves the task without using any helper tools. 

\begin{tcolorbox}[colback=gray!10, colframe=white, title=Example]
The response of Tool Master Agent
\small
\begin{alltt}
[
    \{
    "name": "Web_Scraper", 
    "description": "A tool to extract 
    specific information from web pages
    by crawling and parsing their content."
    \}
]
\end{alltt}
\end{tcolorbox}

\subsection{Tool Retrieval / Generation}
After understanding the tool requirements, the system defines whether it needs to be generated from scratch by the Tool generator or retrieved from the tool dataset. The multi-agent  and tool dataset is used to execute  this portion of the framework. The Multi-agent Tool Generator itself is composed of other  agents. These agents (code writer, code executor, and web scraper) ensure the following:
\begin{itemize}
    \item The tool is generated only when no similar tool is available in the dataset, and
    \item Ensure that the generated tool is functioning correctly.
\end{itemize}

\subsubsection{Tool Database}
ATLASS maintains a database of all the tools available in the system. We use a JSON file to store each tool's 'name', 'description', and 'function-name'. Additionally, we maintain Python scripts that contain the actual implementation of each tool. The 'function-name' in the JSON file serves as a reference, allowing us to retrieve the corresponding Python code for each tool.

\subsubsection{Tool Selector}
\begin{figure}[!ht]
    \centering
    \includegraphics[width=0.5\textwidth]{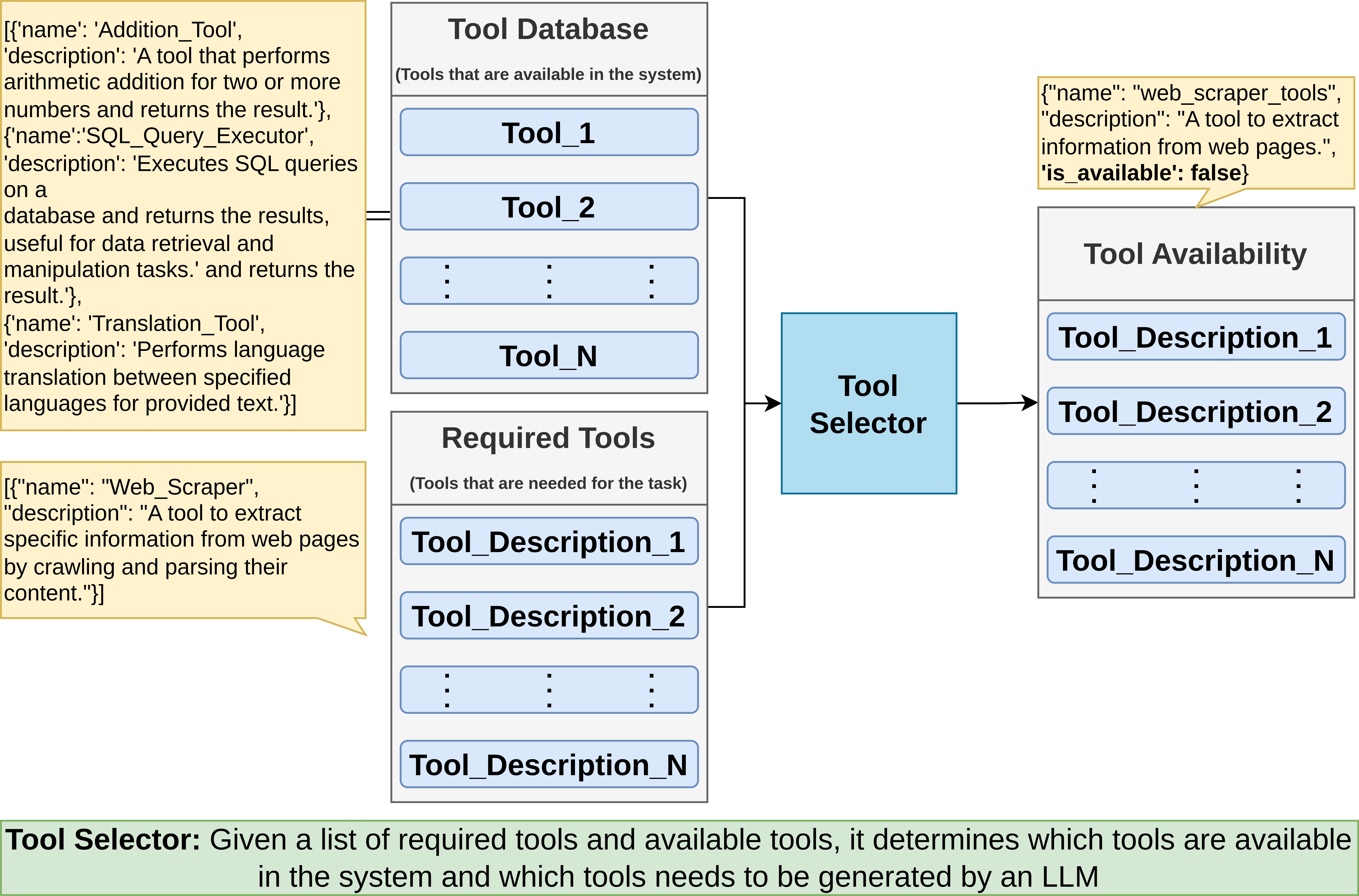}
    \caption{Tool Selector Pipeline}
    \label{fig:toolselector}
\end{figure}

This agent ascertains which necessary tools are already present in the system and which ones require creation using the Tool Dataset and Required Tools to tell us which tools need to be retrieved and which tools need to be generated. Figure~\ref{fig:toolselector} shows how the Tool Selector process works. 

\begin{tcolorbox}[colback=gray!10, colframe=white, title=Example]
The response of Tool Selector
\small
\begin{alltt}
[
    \{
    "name": "web_scraper_tools", 
    "description": "A tool to extract 
    information from web pages.",
    "is_available": false
    \}
]
\end{alltt}
\end{tcolorbox}
\subsubsection{Tool Generator: Non-API-Based}
For the tools that are not available in the system and don't require any kind of API, we use the Tool Generator agent to generate Python code for those tools. The Tool Generator consists of two agents: a code writer and a code executor \ref{fig:toolgenerator} . The process begins by passing the tool's name and description to the code writer. Initially, the code writer generates Python code to install any necessary external packages, ensuring the required dependencies are available for executing the tool's function. We then pass this code to the code executor, essentially a Python interpreter. The code executor installs all the packages and returns the result to the code writer. The code writer ensures that the functions are appropriately annotated. The annotation encompasses the function's purpose, the appropriate usage timing, and comprehensive details such as data type and output description. The code executor then runs that function, and if there are errors, the code executor keeps sending the error back to the code writer. The process operates in a loop, continuously executing until it locates a workable code base. Once it finds a working code base, it adds its information and the tool's information back to the Tool Dataset.

\subsubsection{Tool Generator : API-Based}
For the tools that do require APIs, we at first pass the tool's information to the code writer. When the code writer receives an API-based tool, it simply outputs "API KEY REQUIRED: Name of the API". The reason we don't try to generate API-based tools directly is: 1. An API key is required to run any API-based tool, which the agent currently doesn't have, and 2. API parameters change over time, and it is not necessary that the model knowledge base has the most current information. For these reasons, we use a web scraper agent, which uses a web searching tool (Serp API\cite{serpapiSerpApiGoogle}) to get the current documentation's information on the API usages. The system also asks the user to provide the API key if required. The initial tool information (name and description), the latest documentation content from the internet, and the user-provided API key are combined to create a new prompt, which is then passed to the code generator to generate the tool. 

\begin{figure}[!ht]
    \centering
    \includegraphics[width=0.5\textwidth]{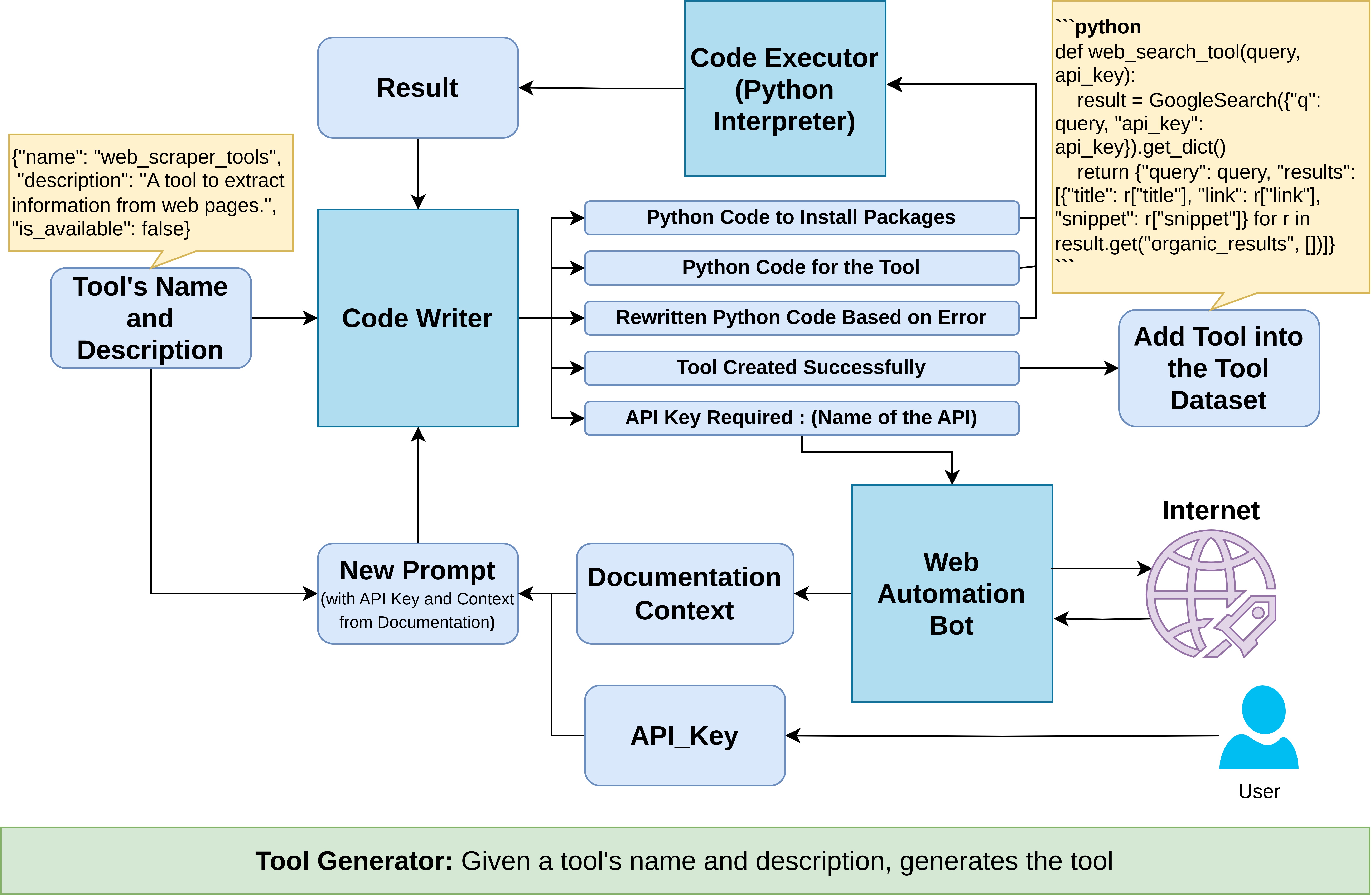}
    \caption{Tool Generator Pipeline}
    \label{fig:toolgenerator}
\end{figure}

Figure~\ref{fig:toolgenerator} shows how the Tool Generator generates the tool. Once required tools are generated or retrieved, the system combines them and passes them to the final stage of the framework.

\begin{tcolorbox}[colback=gray!10, colframe=white, title=Example]
The response of Tool Generator Agent
\small
\begin{alltt}
def web_search_tool(query, api_key):
    result = GoogleSearch({"q": query, 
      "api_key": api_key}).get_dict()
    
    return {"query": query, "results": 
     [{"title": r["title"], "link": 
      r["link"], "snippet": r["snippet"]} 
      for r in result.get("organic_results", [])]}
\end{alltt}
\end{tcolorbox}

\subsection{Solving Tasks}
This final stage of the framework focuses on solving the user's initial query. It's important to mention that if the task does not require any tools, then the information is directly passed to the Task Solver agent from the Tool Master. In such scenarios, the system completely disregards the second part of the framework. If the task requires tools, the system passes the retrieved and generated tools to the Task Solver for task resolution.

\subsubsection{Task Solver}
\begin{figure}[!ht]
    \centering
    \includegraphics[width=0.5\textwidth]{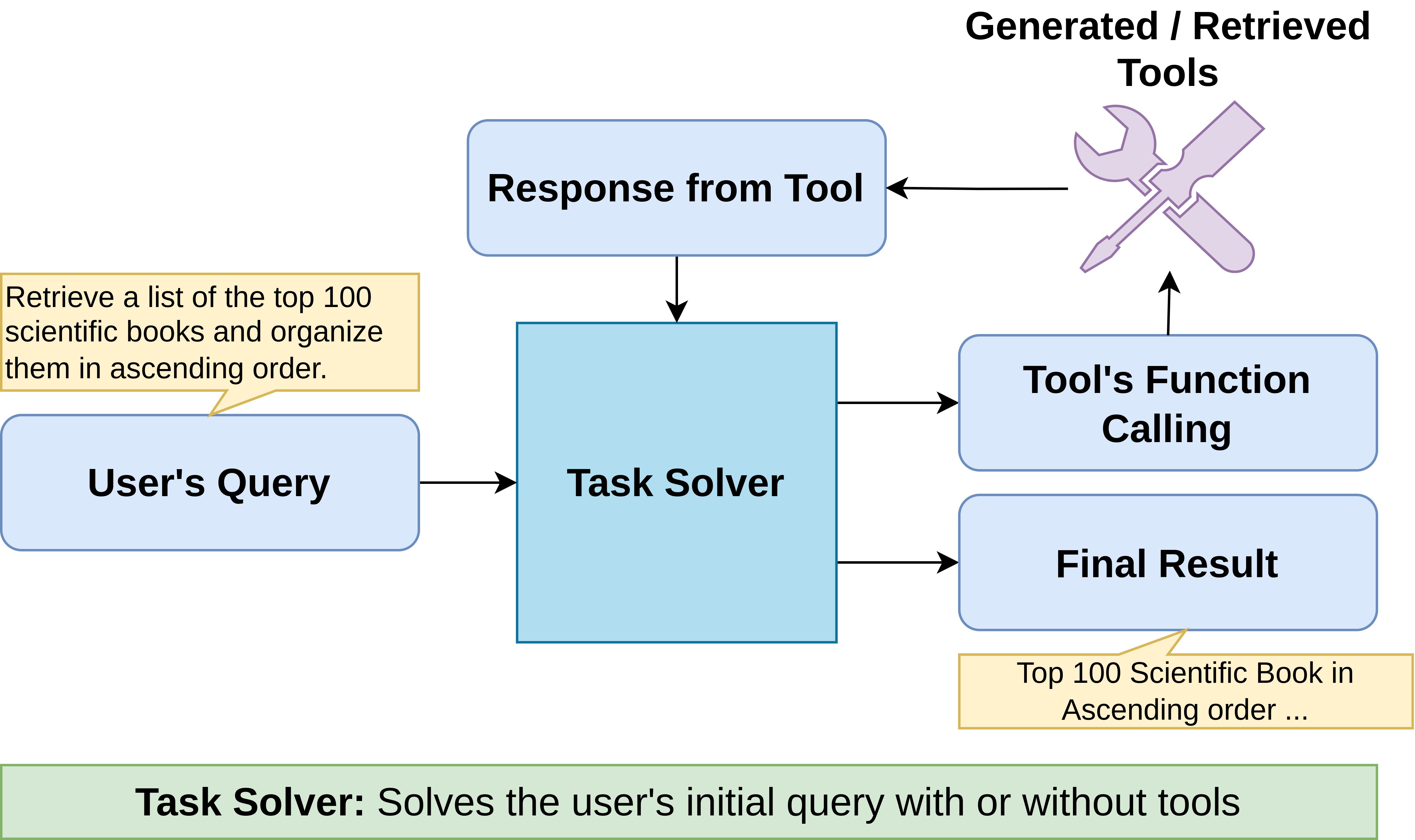}
    \caption{Task Solver Pipeline}
    \label{fig:tasksolver}
\end{figure}

Task Solver is responsible for solving the user's query, with or without external tools. If no tools are required, it relies on its internal knowledge base. Otherwise, it intelligently utilizes the necessary tools to complete the task efficiently. Figure~\ref{fig:tasksolver} shows how Task Solver solves users questions.

\section{Result Analysis}

The performance of the ATLASS framework is influenced by the efficiency of multiple interconnected modules, including the Task Analyzer, Tool Master, Tool Selector, and Tool Generator. As a sequential conversational system, the output of each module directly impacts the subsequent modules, creating a flow throughout the system. We assess multiple criteria to evaluate the overall system performance, taking into account the individual contributions and interdependencies of each module. We evaluate our approach on different domains, including mathematical operations, data analysis, data visualization, forecasting, NLP tasks, and API-based information retrieval.

\subsection{Tool Selection Accuracy}
Given a user prompt, after task analysis and tool requirements analysis, the Tool Selector module defines if the required tool is available in the tool database or not. Let's consider a user prompt: \textit{Generate a bar chart with the last five days stock price of Apple Inc.}, Tool Selector defines that we need two different tools to get the task solved, including \textit{Stock Price Checker} and \textit{Data Visualizer}. Consider we have a tool named \textit{Bar Chart Generator}, Tool Selector module has the ability to understand that the required tool \textit{Data Visualizer} is equivalent to the available tool \textit{Bar Chart Generator}. This ability improves the tool's usability and generalizes its use, as well as reducing the generation of multiple tools for the same task. Table \ref{tab:tool-selector} is a detailed breakdown of how well the Tool Selector performs.

\begin{table}[!ht]
\centering
\renewcommand{\arraystretch}{1.2}
\begin{tabular}{|l|l|}
\hline
\textbf{Tools Name}: & \textit{Word Frequency Counter} \\ \hline
\textbf{Description}:  & \begin{tabular}[c]{@{}l@{}}\textit{A tool that can split a sentence into individual} \\ \textit{words count the frequency of each word, sort} \\ \textit{them according to their frequency, and select} \\ \textit{the most common words.}\end{tabular} \\ \hline
\textbf{Origin Prompt}: & \textit{Find 10 most common words in the sentence.} \\ \hline
\begin{tabular}[c]{@{}l@{}} \textbf{Alternative}  \textbf{Prompts}: \end{tabular} & \begin{tabular}[c]{@{}l@{}}\textit{\textbf{1.} Rank the following keywords in order of } \\ \textit{relevance in this document.} \\ \textit{\textbf{2.} Can you find the unique words in this?} \\ \textit{sentence and tell me how rare they are?}\\ \textit{\textbf{3.} Find out how similar these two texts are} \\  \textit{based on their most commonly used words.}\end{tabular} \\ \hline
\end{tabular}
\caption{The tool \textit{Word Frequency Counter} generated by the prompt \textit{"Find 10 most common words in the sentence."} is reusable by the list of Alternative Prompts}
\label{tab:tool-selector}
\end{table}

\subsection{Tool Generation Performance}
The primary objective of this work is to develop the Tool Generator, which aims to create tools capable of addressing user queries while generalizing prompts to solve similar tasks. We evaluate the performance of our approach in comparison to LATM \cite{cai2024largelanguagemodelstool} across various prompts, highlighting its effectiveness and generalization capabilities. LATM creates tools using language models, generating training, validation, and test sets in response to user prompts. Their pipeline is relatively straightforward: first, a validation set is generated from the user prompt. Next, a Python function is produced based on the prompt to solve the task, with the validation set being used to verify the function's correctness. Finally, the test set corresponds to the original user prompt.

In Table ~\ref{tab:comparison}, the functionalities of ATLASS and LATM are compared. 

\begin{table}[!ht]
\renewcommand{\arraystretch}{1.2}
\begin{tabular}{|c|c|c|@{ }c@{ }|@{ }c@{ }|}
\hline
\textbf{Domain} & \textbf{Task} & \textbf{Tools}   & \textbf{ATLASS} & \textbf{LATM} \\ \hline

\multirow{2}{*}{\begin{tabular}[c]{@{}c@{}}Mathematical \\ Operation\end{tabular}}&Sorting&\textit{sorting-tools}&\ding{51}&\ding{51}\\ \cline{2-5}&Reversing&\textit{string-reverser}&\ding{51}&\ding{51}\\ \hline

\multirow{2}{*}{Data Analysis}&Cleaning&\textit{data-cleaner}&\ding{51}&\ding{51}\\ \cline{2-5}&Extraction&\textit{data-extractor}&\ding{51}&\ding{51}\\ \hline

{Visualization}&\begin{tabular}[c]{@{}c@{}}Graph \\ Generate\end{tabular}&\textit{charting-tools}&\ding{51}&\ding{55}\\ \hline
{Forecasting}&\begin{tabular}[c]{@{}c@{}}Stock \\ Exchange\end{tabular}&\textit{stock-price-checker}&\ding{51}&\ding{55}\\ \hline
{NLP Task}&Sentiment&\textit{sentiment-analyzer}&\ding{51}&\ding{51}\\ \hline
{\begin{tabular}[c]{@{}c@{}}API based \\ Information \\ Extraction\end{tabular}}& SerpAPI&\textit{web-search-tool}&\ding{51}&\ding{55} \\ \hline
\end{tabular}
\caption{Performance Comparison of ATLASS and Language Model as Tool Maker}
\label{tab:comparison}
\end{table}

LATM is capable of generating tools based on user prompts and solving problems using the tools it creates. Since it does not support external library integration, LATM can only generate tools that rely solely on Python's built-in libraries. It also cannot execute tasks that involve API-based information extraction, limiting its use in situations that need data from outside sources.

LATM focuses solely on tool generation, while ATLASS not only generates and saves tools for future use but also use these tools for solving user tasks in a single inference step. This additional focus on reusability and efficient inference significantly enhances the scalability and practicality of the system.

\subsection{Efficiency of Inference}
Because LLM-based agents have a high inference cost, we look into how efficiently ATLASS can solve tasks by generating required tools. At the time of the experiment, we have used the \textit{"gpt-4-0613"} model for all the tasks. On average, we have consumed 2895 tokens and an average cost of 0.1008  USD per prompt when the tool is not available; on the other hand, the consumed tokens are 1920, and the cost is 0.0624 USD per prompt when the tool is available. Table \ref{tab:cost_analysis} shows the token consumption and cost for each prompt. The cost calculation is based on the OpenAI pricing page \cite{pricing}.

\begin{table}[!ht]
\centering
\renewcommand{\arraystretch}{1.2}
\begin{tabular}{|l|ll|ll|}
\hline
\multicolumn{1}{|c|}{\multirow{2}{*}{\textbf{Task}}} & \multicolumn{2}{c|}{\textbf{With no tools in DB}} & \multicolumn{2}{c|}{\textbf{With tools in DB}} \\ \cline{2-5} 
\multicolumn{1}{|c|}{} & \multicolumn{1}{c|}{\textbf{Token}} & \multicolumn{1}{c|}{\textbf{Cost (USD)}} & \multicolumn{1}{c|}{\textbf{Token}} & \multicolumn{1}{c|}{\textbf{Cost (USD)}} \\ \hline
Sorting & \multicolumn{1}{l|}{3161} & 0.1127 & \multicolumn{1}{l|}{2337} & 0.0781 \\ \hline
Reversing & \multicolumn{1}{l|}{3195} & 0.1078 & \multicolumn{1}{l|}{1678} & 0.0536 \\ \hline
Cleaning & \multicolumn{1}{l|}{2627} & 0.0918 & \multicolumn{1}{l|}{1901} & 0.0620 \\ \hline
Extraction & \multicolumn{1}{l|}{2620} & 0.0891 & \multicolumn{1}{l|}{1930} & 0.0617 \\ \hline
Graph Generation & \multicolumn{1}{l|}{3881} & 0.1363 & \multicolumn{1}{l|}{2118} & 0.0703 \\ \hline
Stock Exchange & \multicolumn{1}{l|}{2495} & 0.0875 & \multicolumn{1}{l|}{1740} & 0.0558 \\ \hline
Sentiment & \multicolumn{1}{l|}{2588} & 0.0909 & \multicolumn{1}{l|}{1822} & 0.0586 \\ \hline
SerpAPI & \multicolumn{1}{l|}{2596} & 0.0909 & \multicolumn{1}{l|}{1838} & 0.0591 \\ \hline
\end{tabular}
\caption{Cost Analysis for end-to-end framework}
\label{tab:cost_analysis}
\end{table}

As an end-to-end framework, every query undergoes task analysis, retrieval or generation, and task execution, ensuring the system can handle any given task. The overall cost is minimized by utilizing smaller model variations, such as GPT-3.5.

\section{Conclusion}
This paper presents ATLASS, a closed-loop framework for an advanced tool learning and selection system capable of generating tools automatically for solving complex tasks. ATLAS uses a three-phase approach comprising task analysis and breakdown, tool definition and selection (including retrieval or generation), and task resolution. This approach enables LLM Agents to efficiently solve complex tasks. The framework facilitates the tools to reuse and features a tool selector agent to identify relevant tools for similar problems, ensuring efficiency and adaptability. We demonstrate the ability to create tools that integrate external libraries and API keys for web-based information extraction. Additionally, the framework can seamlessly request user inputs when necessary to provide comprehensive solutions. The system results indicate that the proposed pipeline does a better job of solving complex problems than similar research, demonstrating its superior abilities and usefulness in practice.

\section{Limitations and Future Work}
While ATLASS demonstrates superior tool-generation capabilities compared to previous works, certain limitations remain. This system currently relies on a single model (OpenAI GPT-4.0) for agent testing, which may constrain its adaptability across diverse scenarios. Additionally, challenges persist in generating API-based and highly complex tools, limiting the system’s applicability to certain domains.

To address these limitations, we propose the following directions for future research:
\begin{itemize}
    \item Improve Tool Generator's performance for more complex and API-based tools.
    \item Make the framework more secure by considering possible risks and ethical concerns. Specifically, we want to improve the security of the API keys provided by the user.
    \item Quantitative evaluation of the framework's Tool Generation capability with the comparison of other similar pipelines.     
    \item Apply other models like ChatGPT 3.5 or open-source LLM for less demanding tasks like Task Analyzer and Tool Selector and compare results.
\end{itemize}

\section{Safety and Ethics}

Automating tool generation raises safety, ethical, and security risks, including harmful outputs and exposing API keys. To ensure the ethical and safety aspects of the framework, we have used these criteria:

\begin{enumerate}
    \item \textbf{Human Feedback}
    Generated code poses a security risk \cite{veracode2024risks}. The framework requires human feedback before running any Python code generated by the model, ensuring that a human can read and understand the generated code before execution.
    \item \textbf{Security of API Key}
    API keys can be a security risk when passed directly to a model \cite{trendmicro_chatgpt_security_vulnerabilities}. We are using ChatGPT, so our API key is directly connected to OpenAI. To prevent this, we use regular expressions to insert the API key directly into the code base instead of passing it to the model.
\end{enumerate}

\bibliographystyle{plain} 
\bibliography{bibliography}

\begin{thebibliography}{10}

\bibitem{ahmed2024studyingllmperformanceclosed}
Toufique Ahmed, Christian Bird, Premkumar Devanbu, and Saikat Chakraborty.
\newblock Studying llm performance on closed- and open-source data, 2024.

\bibitem{brown2020languagemodelsfewshotlearners}
Tom~B. Brown, Benjamin Mann, Nick Ryder, Melanie Subbiah, Jared Kaplan, Prafulla Dhariwal, Arvind Neelakantan, Pranav Shyam, Girish Sastry, Amanda Askell, Sandhini Agarwal, Ariel Herbert-Voss, Gretchen Krueger, Tom Henighan, Rewon Child, Aditya Ramesh, Daniel~M. Ziegler, Jeffrey Wu, Clemens Winter, Christopher Hesse, Mark Chen, Eric Sigler, Mateusz Litwin, Scott Gray, Benjamin Chess, Jack Clark, Christopher Berner, Sam McCandlish, Alec Radford, Ilya Sutskever, and Dario Amodei.
\newblock Language models are few-shot learners, 2020.

\bibitem{cai2024largelanguagemodelstool}
Tianle Cai, Xuezhi Wang, Tengyu Ma, Xinyun Chen, and Denny Zhou.
\newblock Large language models as tool makers, 2024.

\bibitem{10.1145/3641289}
Yupeng Chang, Xu~Wang, Jindong Wang, Yuan Wu, Linyi Yang, Kaijie Zhu, Hao Chen, Xiaoyuan Yi, Cunxiang Wang, Yidong Wang, Wei Ye, Yue Zhang, Yi~Chang, Philip~S. Yu, Qiang Yang, and Xing Xie.
\newblock A survey on evaluation of large language models, March 2024.

\bibitem{chen2023autoagents}
Guangyao Chen, Siwei Dong, Yu~Shu, Ge~Zhang, Jaward Sesay, B{\"o}rje~F Karlsson, Jie Fu, and Yemin Shi.
\newblock Autoagents: A framework for automatic agent generation.
\newblock {\em arXiv preprint arXiv:2309.17288}, 2023.

\bibitem{duetting2025multi}
Paul Duetting, Tomer Ezra, Michal Feldman, and Thomas Kesselheim.
\newblock Multi-agent combinatorial contracts.
\newblock In {\em Proceedings of the 2025 Annual ACM-SIAM Symposium on Discrete Algorithms (SODA)}, pages 1857--1891. SIAM, 2025.

\bibitem{gao2023palprogramaidedlanguagemodels}
Luyu Gao, Aman Madaan, Shuyan Zhou, Uri Alon, Pengfei Liu, Yiming Yang, Jamie Callan, and Graham Neubig.
\newblock Pal: Program-aided language models, 2023.

\bibitem{guo2024large}
Taicheng Guo, Xiuying Chen, Yaqi Wang, Ruidi Chang, Shichao Pei, Nitesh~V Chawla, Olaf Wiest, and Xiangliang Zhang.
\newblock Large language model based multi-agents: A survey of progress and challenges.
\newblock {\em arXiv preprint arXiv:2402.01680}, 2024.

\bibitem{hu2022lora}
Edward~J Hu, yelong shen, Phillip Wallis, Zeyuan Allen-Zhu, Yuanzhi Li, Shean Wang, Lu~Wang, and Weizhu Chen.
\newblock Lo{RA}: Low-rank adaptation of large language models.
\newblock In {\em International Conference on Learning Representations}, 2022.

\bibitem{hu}
Renjun Hu, Yi~Cheng, Libin Meng, Jiaxin Xia, Yi~Zong, Xing Shi, and Wei Lin.
\newblock Training an llm-as-a-judge model: Pipeline, insights, and practical lessons, 02 2025.

\bibitem{liu2024tuning}
Alisa Liu, Xiaochuang Han, Yizhong Wang, Yulia Tsvetkov, Yejin Choi, and Noah~A. Smith.
\newblock Tuning language models by proxy.
\newblock In {\em First Conference on Language Modeling}, 2024.

\bibitem{marvin2023prompt}
Ggaliwango Marvin, Nakayiza Hellen, Daudi Jjingo, and Joyce Nakatumba-Nabende.
\newblock Prompt engineering in large language models.
\newblock In {\em International conference on data intelligence and cognitive informatics}, pages 387--402. Springer, 2023.

\bibitem{trendmicro_chatgpt_security_vulnerabilities}
Trend Micro.
\newblock Security vulnerabilities of chatgpt-generated code, 2023.
\newblock Accessed: 2025-02-07.

\bibitem{minaee2024largelanguagemodelssurvey}
Shervin Minaee, Tomas Mikolov, Narjes Nikzad, Meysam Chenaghlu, Richard Socher, Xavier Amatriain, and Jianfeng Gao.
\newblock Large language models: A survey, 2024.

\bibitem{nakano2022webgptbrowserassistedquestionansweringhuman}
Reiichiro Nakano, Jacob Hilton, Suchir Balaji, Jeff Wu, Long Ouyang, Christina Kim, Christopher Hesse, Shantanu Jain, Vineet Kosaraju, William Saunders, Xu~Jiang, Karl Cobbe, Tyna Eloundou, Gretchen Krueger, Kevin Button, Matthew Knight, Benjamin Chess, and John Schulman.
\newblock Webgpt: Browser-assisted question-answering with human feedback, 2022.

\bibitem{ni2025toolfactory}
Xinyi Ni, Qiuyang Wang, Yukun Zhang, and Pengyu Hong.
\newblock Toolfactory: Automating tool generation by leveraging llm to understand rest api documentations.
\newblock {\em arXiv preprint arXiv:2501.16945}, 2025.

\bibitem{pricing}
OpenAI.
\newblock {P}ricing - {O}pen{A}{I} {A}{P}{I}.
\newblock \url{https://platform.openai.com/docs/pricing}.
\newblock [Accessed 28-01-2025].

\bibitem{qin-etal-2023-webcpm}
Yujia Qin, Zihan Cai, Dian Jin, Lan Yan, Shihao Liang, Kunlun Zhu, Yankai Lin, Xu~Han, Ning Ding, Huadong Wang, Ruobing Xie, Fanchao Qi, Zhiyuan Liu, Maosong Sun, and Jie Zhou.
\newblock {W}eb{CPM}: Interactive web search for {C}hinese long-form question answering.
\newblock In Anna Rogers, Jordan Boyd-Graber, and Naoaki Okazaki, editors, {\em Proceedings of the 61st Annual Meeting of the Association for Computational Linguistics (Volume 1: Long Papers)}, pages 8968--8988, Toronto, Canada, July 2023. Association for Computational Linguistics.

\bibitem{schick2023toolformerlanguagemodelsteach}
Timo Schick, Jane Dwivedi-Yu, Roberto Dessì, Roberta Raileanu, Maria Lomeli, Luke Zettlemoyer, Nicola Cancedda, and Thomas Scialom.
\newblock Toolformer: Language models can teach themselves to use tools, 2023.

\bibitem{serpapiSerpApiGoogle}
SerpAPI.
\newblock {S}erp{A}pi: {G}oogle {S}earch {A}{P}{I}.
\newblock \url{https://serpapi.com/}.
\newblock [Accessed 03-02-2025].

\bibitem{shi2024learningusetoolscooperative}
Zhengliang Shi, Shen Gao, Xiuyi Chen, Yue Feng, Lingyong Yan, Haibo Shi, Dawei Yin, Pengjie Ren, Suzan Verberne, and Zhaochun Ren.
\newblock Learning to use tools via cooperative and interactive agents, 2024.

\bibitem{shi2024learning}
Zhengliang Shi, Shen Gao, Xiuyi Chen, Yue Feng, Lingyong Yan, Haibo Shi, Dawei Yin, Pengjie Ren, Suzan Verberne, and Zhaochun Ren.
\newblock Learning to use tools via cooperative and interactive agents.
\newblock {\em arXiv preprint arXiv:2403.03031}, 2024.

\bibitem{talebirad2023multiagentcollaborationharnessingpower}
Yashar Talebirad and Amirhossein Nadiri.
\newblock Multi-agent collaboration: Harnessing the power of intelligent llm agents, 2023.

\bibitem{veracode2024risks}
Natalie Tischler.
\newblock The risks of automated code generation and the necessity of ai-powered remediation, 2024.
\newblock Accessed: 2025-02-07.

\bibitem{wang-etal-2023-self-instruct}
Yizhong Wang, Yeganeh Kordi, Swaroop Mishra, Alisa Liu, Noah~A. Smith, Daniel Khashabi, and Hannaneh Hajishirzi.
\newblock Self-instruct: Aligning language models with self-generated instructions.
\newblock In {\em Proceedings of the 61st Annual Meeting of the Association for Computational Linguistics (Volume 1: Long Papers)}, pages 13484--13508, Toronto, Canada, July 2023. Association for Computational Linguistics.

\bibitem{wu}
Junde Wu, Jiayuan Zhu, and Yuyuan Liu.
\newblock Agentic reasoning: Reasoning llms with tools for the deep research, 02 2025.

\bibitem{wu2024avataroptimizingllmagents}
Shirley Wu, Shiyu Zhao, Qian Huang, Kexin Huang, Michihiro Yasunaga, Kaidi Cao, Vassilis~N. Ioannidis, Karthik Subbian, Jure Leskovec, and James Zou.
\newblock Avatar: Optimizing llm agents for tool usage via contrastive reasoning, 2024.

\bibitem{xu2025largereasoningmodelssurvey}
Fengli Xu, Qianyue Hao, Zefang Zong, Jingwei Wang, Yunke Zhang, Jingyi Wang, Xiaochong Lan, Jiahui Gong, Tianjian Ouyang, Fanjin Meng, Chenyang Shao, Yuwei Yan, Qinglong Yang, Yiwen Song, Sijian Ren, Xinyuan Hu, Yu~Li, Jie Feng, Chen Gao, and Yong Li.
\newblock Towards large reasoning models: A survey on scaling llm reasoning capabilities, 2025.

\bibitem{zhang-etal-2024-mm}
Duzhen Zhang, Yahan Yu, Jiahua Dong, Chenxing Li, Dan Su, Chenhui Chu, and Dong Yu.
\newblock {MM}-{LLM}s: Recent advances in {M}ulti{M}odal large language models.
\newblock In {\em Findings of the Association for Computational Linguistics: ACL 2024}, pages 12401--12430, Bangkok, Thailand, August 2024. Association for Computational Linguistics.

\bibitem{zhang2022automatic}
Zhuosheng Zhang, Aston Zhang, Mu~Li, and Alex Smola.
\newblock Automatic chain of thought prompting in large language models.
\newblock {\em arXiv preprint arXiv:2210.03493}, 2022.

\bibitem{zhao2024surveylargelanguagemodels}
Wayne~Xin Zhao, Kun Zhou, Junyi Li, Tianyi Tang, Xiaolei Wang, Yupeng Hou, Yingqian Min, Beichen Zhang, Junjie Zhang, Zican Dong, Yifan Du, Chen Yang, Yushuo Chen, Zhipeng Chen, Jinhao Jiang, Ruiyang Ren, Yifan Li, Xinyu Tang, Zikang Liu, Peiyu Liu, Jian-Yun Nie, and Ji-Rong Wen.
\newblock A survey of large language models, 2024.

\bibitem{zhao2023survey}
Wayne~Xin Zhao, Kun Zhou, Junyi Li, Tianyi Tang, Xiaolei Wang, Yupeng Hou, Yingqian Min, Beichen Zhang, Junjie Zhang, Zican Dong, et~al.
\newblock A survey of large language models.
\newblock {\em arXiv preprint arXiv:2303.18223}, 2023.

\bibitem{zhou}
Han Zhou, Xingchen Wan, Ruoxi Sun, Hamid Palangi, Shariq Iqbal, Ivan Vulić, Anna Korhonen, and Sercan Arık.
\newblock Multi-agent design: Optimizing agents with better prompts and topologies, 02 2025.

\end{thebibliography}

\end{document}